\documentclass{article}

\usepackage{microtype}
\usepackage{graphicx}
\usepackage{subfigure}
\usepackage{booktabs} 

\usepackage{hyperref}

\usepackage[accepted]{icml2021}

\icmltitlerunning{Negative Inner-Loop Learning Rates Learn Universal Features}

\begin{document}

\twocolumn[
\icmltitle{Negative Inner-Loop Learning Rates Learn Universal Features}




\begin{icmlauthorlist}
\icmlauthor{Tom Starshak}{st}

\end{icmlauthorlist}

\icmlaffiliation{st}{Stanford University}

\icmlcorrespondingauthor{Tom Starshak}{starshak@stanford.edu}

\icmlkeywords{Machine Learning, ICML}

\vskip 0.3in
]



\printAffiliationsAndNotice{} 

\begin{abstract}
Recently, meta-learning (learning to learn) methods have received a lot of attention due to their success in few-shot learning. One branch of meta-learning is optimization based meta-learning, where the ability to quickly adapt to new tasks is explicitly optimized for. For example, Model Agnostic Meta-Learning (MAML) consists of two optimization loops: the outer loop learns a meta-initialization of model parameters that is shared across tasks, and the inner loop performs an "adaptation" step on the specific task it is working on. A variant of MAML, Meta-SGD, uses the same two loop structure, but also learns the learning-rate for the adaptation step. 

The success of these methods has lead to research into the mechanism by which they adapt to new tasks. The original MAML formulation has been shown to work by feature reuse, that is the meta-initialization already contains high quality features that are not changed significantly during adaptation. One variant of MAML (Body Only Update in Inner Loop) forces the model to abandon feature reuse and rapidly change features for every task. However, little attention has been paid to how the learned learning-rate of Meta-SGD affects feature reuse. 

In this paper, we study the effect that a learned learning-rate has on the per-task feature representations in Meta-SGD. The learned learning-rate of Meta-SGD often contains negative values. We hypothesize that during the adaptation phase, these negative learning rates push features away from task-specific features and towards task-agnostic features. 

We performed several experiments on the Mini-Imagenet dataset. Two neural networks were trained, one with MAML, and one with Meta-SGD. The feature quality for both models was tested as follows: strip away the linear classification layer, pass labeled and unlabeled samples through this encoder, classify the unlabeled samples according to their nearest neighbor. This process was performed at three times: 1) after fully training and using the meta-initialization parameters; 2) after adaptation on a task, and validated on that task; and 3) after adaptation on a task, and validated on a different task.  The MAML trained model improved on the task it was adapted to, but had worse performance on other tasks. The Meta-SGD trained model was the opposite; it had worse performance on the task it was adapted to, but improved on other tasks. This confirms the hypothesis that Meta-SGD's negative learning rates cause the model to learn task-agnostic features rather than simply adapt to task specific features.
\end{abstract}

\pagebreak

\section{Introduction}
One of the most important problems in machine learning is few-shot learning; tasks for which there is very limited amounts of training data. This is important since training data is often time consuming and expensive to gather, and in some applications the domain of interest changes over time. Efficiently being able to train a model on a small amount of training data helps with both issues. A lot of research has been done on few-shot learning. The dominant paradigms are transfer learning \cite{huh2016makes}; where a model is trained on a much larger, and related, dataset \cite{deng2009imagenet}, then fine-tuned on the relevant data; and meta-learning \cite{vinyals2017matching, li2017metasgd, finn2017modelagnostic, ravi2016optimization}, or "learning to learn", where a model is explicitly trained to adapt to few-shot tasks quickly.

Meta-learning approaches looks at a family of tasks that share some structure. These tasks are broken into training and evaluation sets. The meta-learning algorithm will then attempt to learn a way in which it can quickly learn adapt to new tasks. Meta-learning approaches tend to fall into one of three categories: black-box algorithms, optimization based algorithms, and non-parametric approaches. 

In optimization based approaches \cite{finn2017modelagnostic, li2017metasgd, nichol2018firstorder}, the loss function explicitly takes into account the multi-task aspect of the problem and the model optimizes the ability to quickly adapt to new tasks. A very successful examples of this approach {\it Model Agnostic Meta-Learning (MAML)}. \cite{finn2017modelagnostic} Broadly, MAML consists of two optimization loops. The outer loop iterates over different tasks and learns a meta-initialization such that the inner loop can quickly adapt to new tasks. The inner loop iterates over data points from the same task and optimizes the relevant loss function. A MAML variant, Meta-SGD \cite{li2017metasgd}, uses the same approach but also learns the inner loop learning-rate.

Due to the influence of MAML, a topic of research that has been explored is what exactly is happening in the underlying neural network \cite{raghu2020rapid, goldblum2020unraveling}. This is often framed as a question of {\it rapid learning}, where the feature representations change dramatically between different tasks, or {\it feature reuse}, where the meta-initialization that is learned is already very good and the feature representations are changed only slightly. Vanilla MAML has been shown to work via feature reuse, while some variants \cite{oh2021boil} force the underlying model to rapidly adapt. Other research \cite{bernacchia2021metalearning} has shown that different learning rates can have dramatic effects on MAML's performance. Surprisingly, little research has been done on how Meta-SGD's learned learning-rates affect the feature representations of the underlying model. In this paper we explore this question. Our main contributions are:

\begin{itemize}
    \item Exploration of the distribution of per-parameter learning rates that result from training with Meta-SGD, finding that most of the learning-rate parameters become negative.
    \item Categorizing the effectiveness of the meta-initialization features for both MAML and Meta-SGD, confirming previous work that shows feature reuse is dominant.
    \item Investigate the changes that the feature representations undergo during task-specific adaptation. We find that adaptation for both MAML and Meta-SGD cause slight decrease in performance for that task. However MAML decreases performance on tasks it wasn't adapted to, while Meta-SGD increases performance on tasks it wasn't adapted to.
    \item Discuss why negative learning rates allow Meta-SGD to learn task-agnostic features.
\end{itemize}

\section{Related Work}

This section represents a brief overview of necessary background in the used meta-learning algorithms and feature representation studies.

\subsection{Meta-Learning Algorithms}
\subsubsection{Model-Agnostic Meta-Learning}

Model-Agnostic Meta-Learning \cite{finn2017modelagnostic}is a conceptually simple algorithm that holds great power and can be used with any differentiable model, usually a neural network. Consider a model $\textit{f}_{\theta}$ that maps observations \textbf{x} to observations \textbf{a}. We will consider different tasks $\mathcal{T}$, where the observations are drawn from different distributions. We denote the loss function as $\mathcal{L}$. When adapting to a new task, the parameters of the model $\theta$ become $\theta'_i$ using a gradient update:

\begin{equation}
\theta'_i = \theta - \alpha \nabla_{\theta} \mathcal{L}_{\mathcal{T}_i}(f_{\theta})
\end{equation}

Where $\alpha$ is the inner learning rate. This update is only for one task. Since we want to optimize the ability to learn across tasks, several tasks are sampled, the updates are performed, and a meta-update is performed across the sum of losses for all these tasks.

\begin{equation}
\theta = \theta - \beta \nabla_{\theta} \sum_{\mathcal{T_i}} \mathcal{L}(f_{\theta'_i})
\end{equation}

This algorithm eventually learns parameters such that once a task is sampled a gradient step for that task will result in good performance.

\begin{figure}[ht]
\vskip 0.2in
\begin{center}
\centerline{\includegraphics[width=\columnwidth]{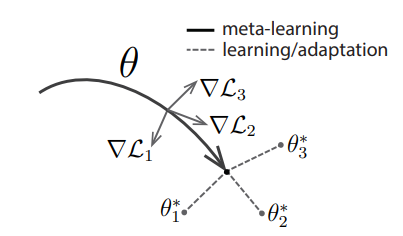}}
\caption{Diagram of the MAML algorithm, showing adaptations to new tasks.}
\label{maml}
\end{center}
\vskip -0.2in
\end{figure}

\subsubsection{Meta-SGD}

Meta-SGD \cite{li2017metasgd} is extremely similar to MAML in that it was an outer-loop to learn a meta-initialization and an inner-loop to adapt to different tasks. The difference between the two algorithms is that Meta-SGD learns per-parameter learing rates for tha adaptation step.  The inner loop update is nearly the same:

\begin{equation}
\theta'_i = \theta - \alpha \circ \nabla_{\theta} \mathcal{L}_{\mathcal{T}_i}(f_{\theta})
\end{equation}

and where the outer-loop must also update the learning rate.

\begin{equation}
\theta = \theta - \beta \nabla_{\theta, \alpha} \sum_{\mathcal{T_i}} \mathcal{L}(f_{\theta'_i})
\end{equation}

This gives Meta-SGD a lot more flexibility in that the step-size and step direction are adapted; the model is not constrained to step in the same direction as the gradient.

\begin{figure}[ht]
\vskip 0.2in
\begin{center}
\centerline{\includegraphics[width=\columnwidth]{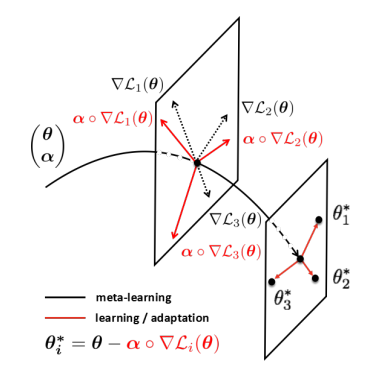}}
\caption{Diagram of the Meta-SGD algorithm. Meta-initialization of the parameters and learning rate allow flexible adaptation.}
\label{meta-sgd}
\end{center}
\vskip -0.2in
\end{figure}

\subsubsection{Prototypical Networks}

Prototypical networks (proto-nets) \cite{snell2017prototypical} differ from the previous two algorithms in that while they update model parameters with optimization schemes, while proto-nets are non-parametric. In a proto-net, an embedding is learned such that embeddings of the same classes are close to each other. The prototype of each class embedding is the mean vector of all the training points of that class.

\begin{equation}
    c_k = \frac{1}{|S_k|} \sum_{S_k} f_{\phi}(x_i)
\end{equation}

Where $S_k$ is the set of all examples of the k-th class and $f_{\phi}$ is the embedding function.

During test time, examples are passed through the embedding function and examples are simply classified according to the nearest prototype.

\begin{figure}[ht]
\vskip 0.2in
\begin{center}
\centerline{\includegraphics[width=\columnwidth]{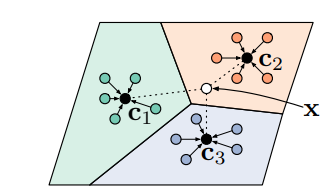}}
\caption{Few-shot prototypical network. The example X would be classified as class $c_2$.}
\label{proto-net}
\end{center}
\vskip -0.2in
\end{figure}

\subsection{Meta-Learning Feature Representations}

There has been a fair amount of research into what sorts of features are learned with MAML-like algorithms. In one study \cite{goldblum2020unraveling} the authors conjecture that the success of meta-learning algorithms suggests that the features that they learn are fundamentally different than classically fine-tuned feature. They show that the feature representations learned through meta-learning are much more tightly clustered than traditional approaches and more easily separable. Another study \cite{raghu2020rapid} considered only meta-learning representations, but looked at how the representations changed from before the adaptation step to after it. They showed that adaptation does not change the representation very much; nearly all the changes in the inner loop occur in the classification head. In fact, they propose an algorithm (Almost No Inner Loop/ANIL) that freezes most of the neural network during adaptation, yet still has nearly the same performance on the MiniImageNet benchmark as regular MAML.

\section{MAML, Meta-SGD, and Universal Features}

Our goal is to understand how the difference between MAML and Meta-SGD, specifically, the negative learning rates that Meta-SGD learns, affect the feature representations of the underlying model. As shown in \ref{maml} and \ref{meta-sgd} the parameters of the model move from a meta-initialization to a new space. In MAML, this is easy to understand. The inner-loop moves down gradient from the meta-initialization because it has a static learning rate. The model then performs better on the adaptation task. What happens in Meta-SGD is less clear. Often, large parts of the learning rate vector are negative. During adaptation some parameters are moved away from what would improve the model's performance on the adaptation task. That negative learning rates are still learned implies that this is beneficial. We conjecture that the purpose of these negative learning rates is to improve features for other tasks than the current one. That is, negative learning rates cause the model to learn universal features.

\section{Experiments}

\begin{figure}[ht]
\vskip 0.2in
\begin{center}
\centerline{\includegraphics[width=\columnwidth]{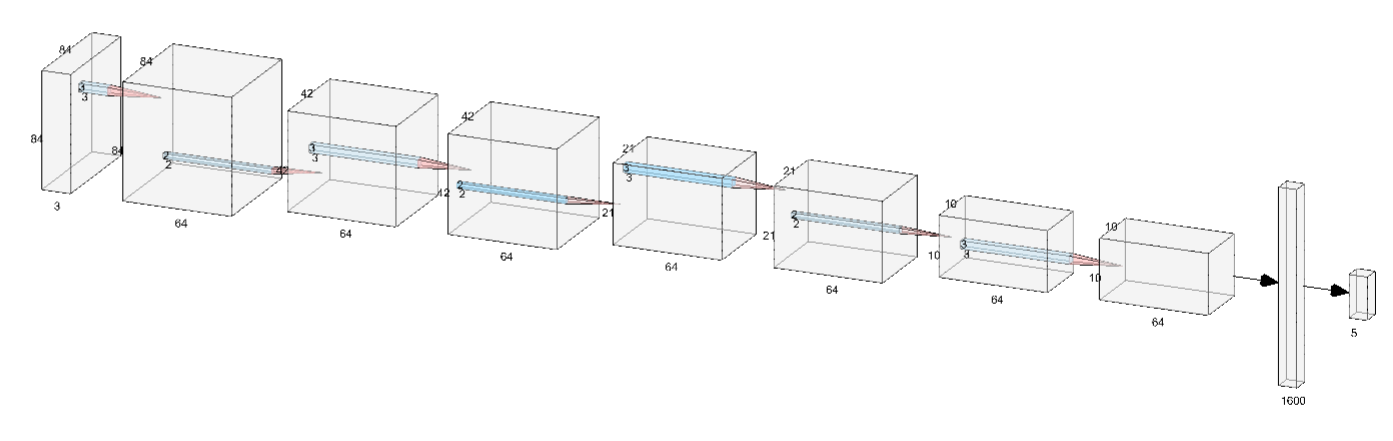}}
\caption{The convolutional neural network that was trained with both MAML and Meta-SGD.}
\label{model}
\end{center}
\vskip -0.2in
\end{figure}

All experiments require two models, both using the same architecture as shown in \ref{model}, be trained for image classification on Mini-ImageNet \cite{vinyals2017matching}. The model has four convolutional blocks which each consist of: 3x3 convolution, BatchNorm, ReLU activation, and MaxPool. After the four convolutional blocks, the representation is flattened and fed into a linear layer. The loss function is the cross-entropy loss.

\begin{algorithm}[tb]
   \caption{Meta-Learning algorithms. If $\alpha$ is learnable this is Meta-SGD, otherwise it is MAML}
   \label{meta-algo}
\begin{algorithmic}
   \STATE {\bfseries Require:} $p(\mathcal{T})$: distribution over tasks
   \STATE {\bfseries Require:} $(\alpha), \beta$: step size parameters
   \STATE Initialize $\theta$
   \WHILE{not done}
   \STATE Sample $\mathcal{T}_i~p(\mathcal{T})$
   \FOR{all $\mathcal{T}_i$}
   \STATE Evaluate $\nabla_{\theta}\mathcal{L}_{\mathcal{T}_i}(f_{\theta})$
   \STATE $\theta'_i = \theta - \alpha \nabla_{\theta}\mathcal{L}_{\mathcal{T}_i}(f_{\theta})$
   \ENDFOR
   \STATE $\theta \gets \theta - \beta \nabla_{\theta, (\alpha)} \sum_{\mathcal{T}_i}\mathcal{L}_{\mathcal{T}_i}(f_{\theta}) $
   \ENDWHILE
\end{algorithmic}
\end{algorithm}

One model each was trained with MAML and Meta-SGD as shown in \ref{meta-algo}. The models were trained as 5-way, 1-shot models. The inner-loop sampled 3 tasks per update and 5 gradient steps per task. For MAML, the inner-loop learning rate was 0.01. In the outer-loop, 10 samples per class were used to calculate the meta-gradient. Both models were trained for 60,000 iterations on an NVIDIA GTX 1070 video card.

After training the meta-initializations were no longer updated. To examine feature reuse, embeddings were created three times: before the inner-loop adaptation, after inner-loop adaptation, and for the task the model was adapted to, and after adaptation for a different task. These will be refered to as "pre-adaptation", "on-task", and "off-task." 

We want to concentrate on the feature embeddings that are created by a model and not the classification of those embeddings. To create an embedding, the linear layer was removed from the neural network and an example image fed through the convolutional layers. This results in a 1600-dimensional vector. One example for each class in a task was used as the centroid of a protonet. For inference, an evaluation image is simply classified as the class with the most similar embedding when evaluated by cosine similarity. A single iteration consisted of: a pre-adaptation evaluation, sampling a task, adapting the model to that task with 5 gradient steps and the appropriate inner-loop learning rate, on-task evaluation, sampling a different task, and an off-task evaluation. The models were evaluated for 40,000 iterations.

\section{Results}
\subsection{Base Training}

The MAML and Meta-SGD base models finished training with validation accuracies of approximately 46\% and 47\% respectively. These are both a bit below what was achieved in the MAML and Meta-SGD studies (48.7\% and 50.5\%) likely owing to the different meta-batch sizes used.

\begin{figure}[ht]
\vskip 0.2in
\begin{center}
\centerline{\includegraphics[width=\columnwidth]{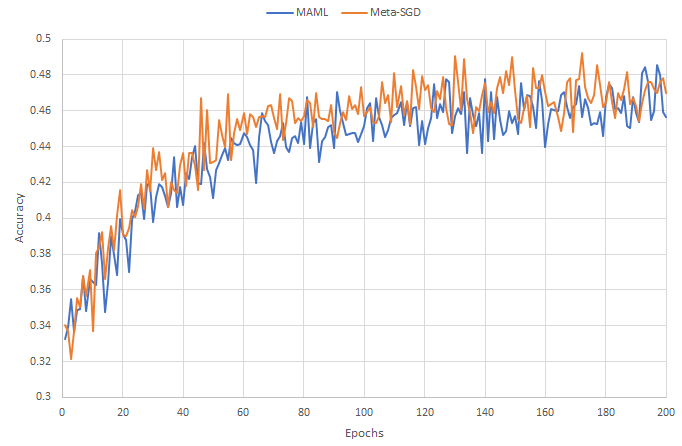}}
\caption{5-way, 1-shot training accuracies on Mini-ImageNet}
\label{training_results}
\end{center}
\vskip -0.2in
\end{figure}

\subsection{Meta-SGD Learning-Rates}

\begin{figure}[ht]
\vskip 0.2in
\begin{center}
\centerline{\includegraphics[width=\columnwidth]{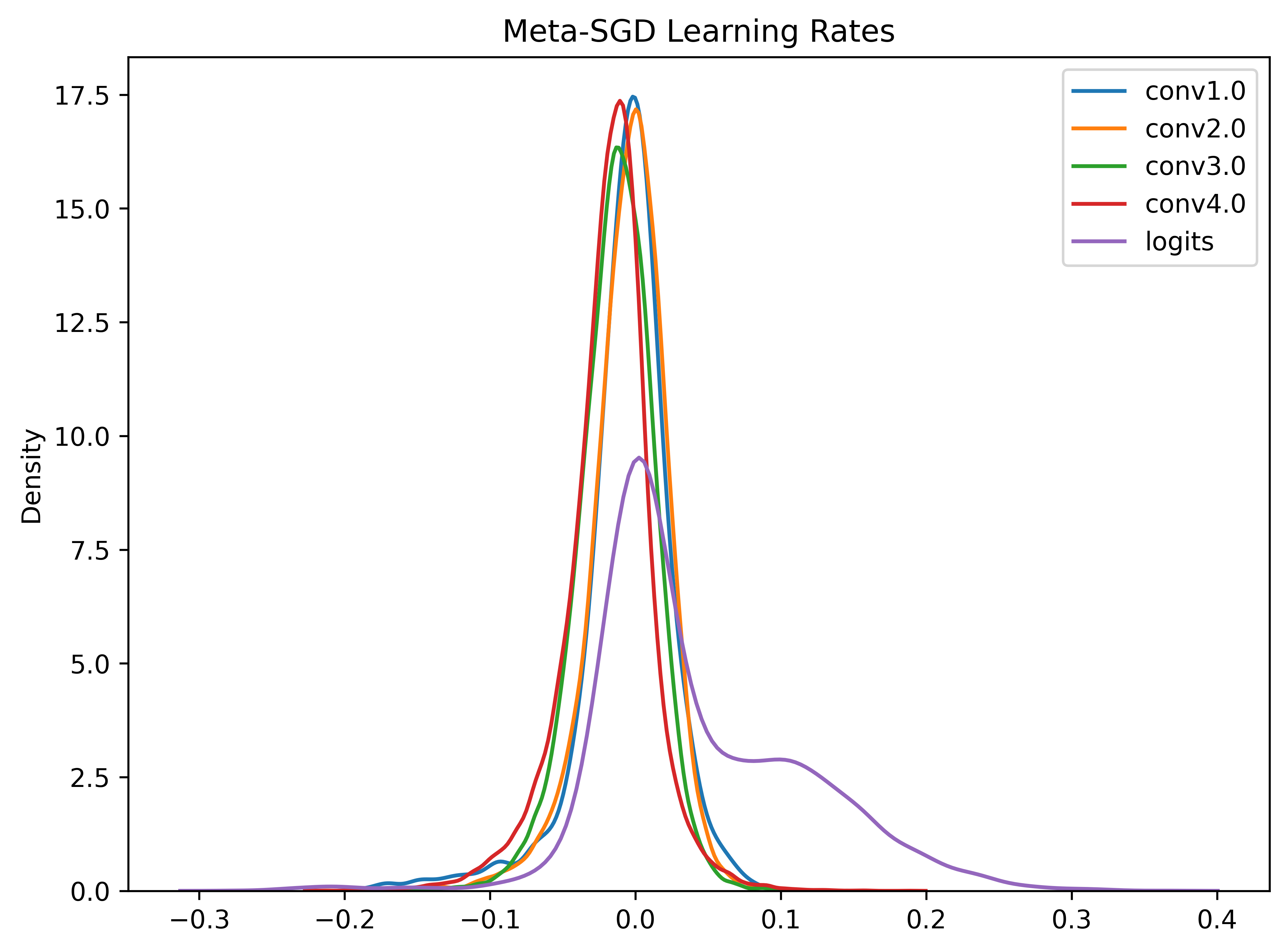}}
\caption{The learned inner-loop learning rates for the Meta-SGD model.}
\label{meta-sgd lr}
\end{center}
\vskip -0.2in
\end{figure}

The inner-loop learning rate after 60,000 iterations can be seen in ~\ref{meta-sgd lr} and the distribution statistics for each layer is shown in \ref{mean lrs}. It is interesting to note that the mean learning rate for every convolutional layer is negative and only the mean learning rate for the linear output layer is positive. 

\begin{table}[t]
\caption{Final inner-loop learning rates for meta-SGD.}
\label{mean lrs}
\vskip 0.15in
\begin{center}
\begin{small}
\begin{sc}
\begin{tabular}{lcccr}
\toprule
Layer & Mean & Std. Dev.\\
\midrule
Conv1    & -0.006 & 0.032 \\
Conv2    & -0.004 & 0.027\\
Conv3    & -0.013 & 0.027\\
Conv4    & -0.018  & 0.030     \\
Logits   & 0.043 & 0.072\\
\bottomrule
\end{tabular}
\end{sc}
\end{small}
\end{center}
\vskip -0.1in
\end{table}

\subsection{Feature Adaptation}

\begin{figure}[ht]
\vskip 0.2in
\begin{center}
\centerline{\includegraphics[width=\columnwidth]{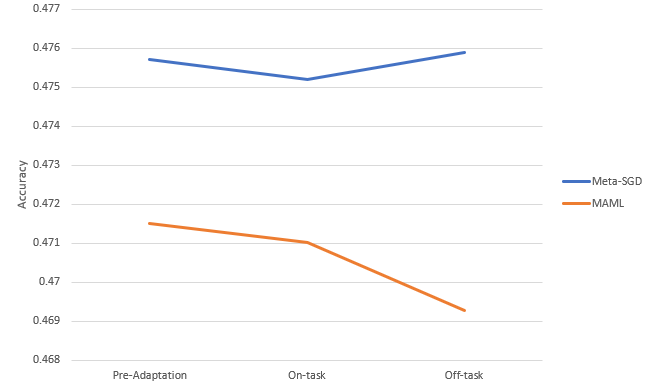}}
\caption{Average accuracy for proto-net based evaluation on embedding vectors for both MAML and Meta-SGD neural networks. Pre-adaptation is using the meta-initialization. On-task is evaluated on the same task that the network was adapted to. Off-task is evaluated on a different task than the model was adapted to.}
\label{final acc}
\end{center}
\vskip -0.2in
\end{figure}

Numerical results from 40,000 iterations of pre-adaptation, on-task, and off-task evaluation can bee seen in \ref{final acc}.

There are a variety of very interesting things to note here. First of all, both models achieve accuracy that is nearly as good (Meta-SGD) or slightly better (MAML) than they achieved using the regular evaluation procedure (Note that the accuracy reported in \ref{final acc} is averaged over 40,000 iterations while the accuracy reported in \ref{training_results} were only averaged over 40 iterations). That is, the features that are learned in the meta-initialization, \textit{without a classification layer or adaptation} retain almost all predictive performance. This is more evidence for the conclusions reached in \cite{raghu2020rapid}, that the function of MAML is feature re-use and not rapid learning. 

Again, we can also see that Meta-SGD outperforms MAML in this setting as well. This shows that the extra flexibility, in both step-size and step-direction, that are present in Meta-SGD serve to learn better feature representations than MAML. It is not merely that Meta-SGD allows the model to take more productive steps during adaptation.

A surprising result is that both models perform worse in the on-task regime.This can be explained for Meta-SGD; in \ref{meta-sgd lr} we see that the mean learning-rate for all of the convolutional layers are negative. Since these are the layers that contribute to the embedding ,the adaptation should then push the weights away from what would improve the model for that task. This is not the case for MAML, where there is a single positive weight for all parameters. This indicates that the classification layer is significant for MAML during the adaptation step. 

The final and most interesting result is what occurs in the off-task regime. MAML trained models degrade the representations of features that are not present in the current task while Meta-SGD trained models improve the representations of features that are not present in the current task. The Meta-SGD trained model increased in accuracy by ~0.2\% while the MAML trained model decreased in accuracy by ~0.3\%. While these results are modest, a t-test comparing the relative changes between on-task and off-task performance shows a p-value = 0.00044, indicating that this is a real phenomenon.

Meta-SGD learning better feature representations in the off-task regime suggests it may learn in a novel way. The negative learning rates push weights away from on-task representations and therefor toward off-task representations rather than how MAML works.

\section{Future Work}
We have shown that when trained on Mini-ImageNet, the Meta-SGD algorithm will learn negative inner-loop learning rates and that these learning rates push parameters away from on-task feature representation and towards off-task (universal) feature representations. 

Several potential areas of inquiry were not explored due to time constraints:

\begin{itemize}
    \item Does this result hold across different datasets?
    \item How does the number of inner-loop gradient updates affect the learned inner-loop learning rate in Meta-SGD? How does it affect the representation change?
    \item Meta-SGD improving performance on off-task representation suggests there may be a better way to evaluate an example at test time, by either not adapting or by adapting to a different task (or tasks) than those that are of interest.
    \item What mechanism causes the representation learned by MAML to perform worse in the on-task regime?
\end{itemize}

\section{Contributions}

Tom Starshak was the sole contributor for this report. Base code for the MAML implementation was taken and adapted from: \hyperlink{https://github.com/oscarknagg/few-shot}{here}.

\bibliography{main.bib}
\bibliographystyle{icml2021}

\end{document}